\documentclass[letterpaper]{article} 
\usepackage{aaai2026}  
\usepackage{times}  
\usepackage{helvet}  
\usepackage{courier}  
\usepackage[hyphens]{url}  
\usepackage{graphicx} 
\usepackage{natbib}  
\usepackage{caption} 
\usepackage{algorithm}
\usepackage{algorithmic}
\usepackage{booktabs}       
\usepackage{amsmath,amssymb}
\usepackage{cleveref}
\usepackage{multirow}
\usepackage{newfloat}
\usepackage{listings}
\DeclareCaptionStyle{ruled}{labelfont=normalfont,labelsep=colon,strut=off} 
\floatstyle{ruled}
\newfloat{listing}{tb}{lst}{}
\floatname{listing}{Listing}
\title{PocketLLM: Ultimate Compression of Large Language Models via Meta Networks}
\author{
	Ye Tian\textsuperscript{\rm 1}, 
	Chengcheng Wang\textsuperscript{\rm 2}, 
	Jing Han\textsuperscript{\rm 3}, 
	Yehui Tang\textsuperscript{\rm 1},
	Kai Han\footnote{Corresponding author}\textsuperscript{\rm 1}
}
\affiliations{
	\textsuperscript{\rm 1} Huawei Noah’s Ark Lab\\
	\textsuperscript{\rm 2} University of Sydney\\
	\textsuperscript{\rm 3} School of Artificial Intelligence, Beijing University of Posts and Telecommunications\\
	\{tianye76, kai.han\}@huawei.com, hanj@bupt.edu.cn
}

\usepackage{bibentry}

\begin{document}

\maketitle

\begin{abstract}
	As Large Language Models (LLMs) continue to grow in size, storing and transmitting them on edge devices becomes increasingly challenging. Traditional methods like quantization and pruning struggle to achieve extreme compression of LLMs without sacrificing accuracy. In this paper, we introduce PocketLLM, a novel approach to compress LLMs in a latent space via meta-networks. A simple encoder network is proposed to project the weights of LLMs into discrete latent vectors, which are then represented using a compact codebook. A lightweight decoder network is employed to map the codebook's representative vectors back to the original weight space. This method allows for significant compression of the large weights in LLMs, consisting solely of a small decoder, a concise codebook, and an index. Extensive experiments show that PocketLLM achieves superior performance even at significantly high compression ratios, e.g., compressing Llama 2-7B by 10$\times$ with a negligible drop in accuracy.
\end{abstract}


\section{Introduction}
The rapid development of large language models (LLMs) ~\cite{openai2023chatgpt,touvron2023llama,zeng2023glm-130b} has been one of the most impressive advancements in artificial intelligence in recent years. These models have grown larger and larger in scale, with their capacity and capabilities advancing at an unprecedented rate, following the scaling law of pretraining~\cite{kaplan2020scaling}. As a result, they have demonstrated remarkable proficiency across a wide range of tasks, including natural language understanding and generation to specialized applications in fields such as personal assistant, finance, education, and so on~\cite{openai2023chatgpt,yang2023fingpt,li2025gumiho}.

The increasing scale of these models has not only pushed the boundaries of what is possible but has also raised significant challenges related to efficiency and practical deployment. More and more mobile devices tend to integrate the capabilities of LLMs, such as laptops, smartphones, and autonomous vehicles~\cite{gunter2024apple,xu2024drivegpt4}. However, these devices are often limited in terms of storage space, and the large size of large models makes them impractical for direct deployment. Additionally, the transmission and regular updating of these models from cloud to devices require substantial network bandwidth, which may not always be available or cost-effective, especially in regions with limited connectivity, seriously affecting the user experience. Consequently, there is a pressing need for extreme compression techniques that can significantly reduce the model's size while preserving its performance, enabling efficient storage, faster transmission, and seamless deployment on resource-constrained devices.

Recent studies have shown that traditional methods including pruning and quantization can preserve the performance of LLMs at a low compression ratio~\cite{kurtic2023sparse,wei-etal-2024-structured,dettmers2023spqr,guo2024compressing,liao2025spark,huang2025sola}. However, as the compression ratio continues to increase, there will be a significant decrease in accuracy. As the compression ratio increases, pruning and low-bitwidth quantization methods unavoidably lose some crucial information, thereby affecting accuracy. Meanwhile, these methods often require a large amount of computing expenditure to retrain or finetune the LLMs. Though some post-training methods are proposed to compress LLMs without retraining, these methods mostly quantify LLMs with 3-4 bitwidth ~\cite{frantar2022gptq,OmniQuant,lin2024duquant,chee2023quip,kim2023squeezellm}. With the proposal of Low-Rank Adaptation (LoRA)~\cite{hu2022lora}, some methods introduce additional trainable parameters after compression to ensure accuracy~\cite{ji-etal-2024-adaptive,xu2023qaloraquantizationawarelowrankadaptation,wang2025SLMQuant,ashkboos2024slicegptcompresslargelanguage}. Nevertheless, the accuracy is still limited at extreme compression ratios (e.g., $>10\times$), and the complex fine-tuning process also complicates the pipelines.

In this paper, we propose to compress the large language models in latent spaces, instead of directly quantizing or pruning the weight matrix as it is. In particular, we first view the weight matrix as a collection of multiple discrete short vectors given that it is challenging to perform quantization or pruning in units of the entire weight matrix. Considering that the complex relationship and redundancy of weight vectors are difficult to learn in the original linear vector space, we construct a meta encoder network to project the weight vectors into latent vectors. We learn a small codebook to represent these latent vectors discretely. We also construct a meta decoder network to map the discrete vectors back to the original linear space for the weight reconstruction. In this way, we only need to store the meta decoder networks, the codebook, and the corresponding indices of latent vectors, so as to compress the weights of LLMs extremely at the cost of only a few additional parameters. Extensive experiments across various popular benchmarks show that our method achieves new state-of-the-art results at extreme compression ratios. Even at $16\times$ compression, the performance loss of PocketLLM with fine-tuning is still acceptable.

\section{Related Works}

\textbf{Traditional Pruning Methods} Traditional pruning methods fall broadly into two categories, i.e. structured pruning and unstructured pruning. For structured pruning methods, researchers often remove channels or layers with non-critical information. SoBP~\cite{wei-etal-2024-structured} selects the pruning structures based on global first-order information and leverages a local greedy approach to refine them. To mitigate information loss, they adopt module-wise reconstruction as an alternative to fine-tuning. LLM-Pruner~\cite{ma2023llmpruner} selectively removes non-critical coupled structures based on gradient information and utilizes Low-Rank Adaptation(LoRA) fine-tuning to recover the performance of pruned models. For unstructured pruning, they zero out the weights of multiple neurons without modifying the model structure. SparseGPT~\cite{frantar-sparsegpt} focuses on the extremely large-scale instances of sparse regression and proposes an efficient one-shot pruning method. Wanda~\cite{sun2023wanda} evaluates the priority of each weight by the product of its magnitude and the norm of the corresponding input feature. For each output of linear layers, the weights with lower priority are pruned. 

\textbf{Traditional Quantization Methods} Traditional quantization methods can be roughly divided into two categories, i.e. quantization during training and post-training methods. Due to the fact that the former methods require large computational overhead to retrain LLMs during the quantification process, academia focuses more on the study of the latter. GPTQ~\cite{frantar2022gptq} proposes an efficient post-training method that minimizes layer-wise quantitative loss based on approximate second-order information. ZeroQuant~\cite{yao2022zeroquant} proposes a fine-grained hardware-friendly quantization scheme that makes the group-wise quantization for weights and token-wise quantization for activation, respectively. OmniQuant~\cite{OmniQuant} proposes an efficient quantization framework that optimizes both weight-only and weight-activation quantization through block-wise error minimization. 

%
\textbf{Codebook-based Methods} Codebook-based methods also become a research hotspot in recent years~\cite{quip-sharp,liu-etal-2024-vptq,NEURIPS2024_6de2e84b,baalen2024gptvq,egiazarian2024extreme}. The concept of codebook is usually applied in the field of image generation. To compress a database of image vectors or embeddings, researchers learn a relatively small set of discrete vectors called "codebook" and optimize these vectors using nearest neighbor algorithms. For each given database vector, we can choose the nearest vector in the codebook for representation. This compression process is also known as vector quantization (VQ). For example, AQLM~\cite{egiazarian2024extreme} constructs an end-to-end quantization method based on multi-group codebooks. It combines vectors from multiple codebooks to represent a weight vector, thereby reducing the difficulty of approximation. VPTQ~\cite{liu-etal-2024-vptq} utilizes Second Order Optimization with brief codebooks to refine the post-training quantization for LLM. QTIP~\cite{NEURIPS2024_6de2e84b} instead uses trellis coded quantization to separate the codebook size from the bitrate and dimension to improve the performance further. Although compared to traditional methods, the compression rate is significantly improved while the accuracy remains better. However, these methods need to construct complex fine-tuning pipelines to restore the accuracy of the compressed models. Essentially, it is because of the high difficulty of constructing codebooks with good representational ability in the current linear space.
\begin{figure*}[htp]
	\centering
	\includegraphics[width=0.95\linewidth]{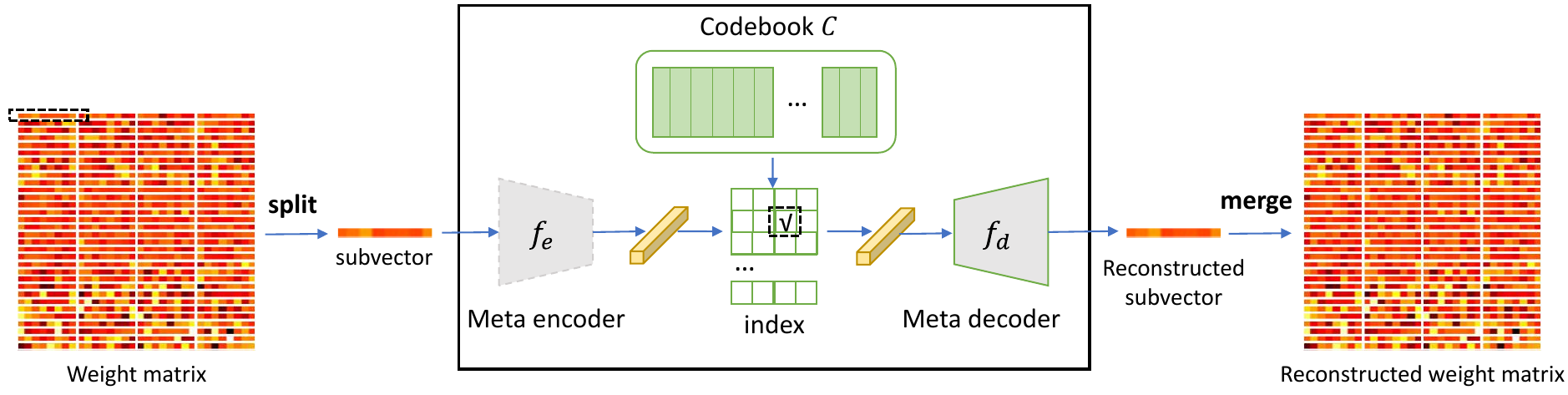}
	\caption{PocketLLM framework: compressing weight vectors in latent spaces with meta-networks.}
	\label{fig:main}
	\vspace{-1em}
\end{figure*}
\section{Approach}
\subsection{Prelimineries}
\paragraph{Transformer}
Transformer is composed of interleavedly stacked multi-head self-attention modules and feed-forward network blocks. The self-attention performs as follows:
\begin{equation}
	O = W_o\textit{Softmax}(QK^T/\sqrt{d})V,
\end{equation}
where $W_o$ is the output projection matrix, $d$ is the feature dimension of queries, $Q$, $K$ and $V$ are obtained by fully-connected layers with weights $W_q$, $W_k$ and $W_v$ from the input $X$ respectively:
\begin{align}
	Q &= W_qX,\\
	K &= W_kX,\\
	V &= W_vX.
\end{align}
The number of parameters in the self-attention module is about $4d^2$.

Most parameters of the feed-forward network are occupied by the two fully-connected layers:
\begin{equation}
	X' = W_{down}\sigma(W_{up}X),
\end{equation}
where $W_{up}$ and $W_{down}$ are the weight matrices, and $\sigma$ is the nonlinear activation function. The hidden dimension is usually set larger than the input dimension, \emph{e.g.}, $4d$. The number of parameters in the FFN module is about $8d^2$. Some variants of the standard FFN have more fully-connected layers like Llama with $12d^2$ parameters.

\subsection{PocketLLM Framework}
As Transformers in LLMs continue to scale, storing and transmitting them on edge devices becomes increasingly challenging. Traditional methods such as quantization and pruning often fail to achieve extreme compression without sacrificing accuracy. In this paper, we introduce PocketLLM for compressing LLMs in latent spaces using meta-networks. The entire pipeline is illustrated in Figure~\ref{fig:main}. First, we divide the weight matrix into multiple vectors and use an encoder network to project these weight vectors into the latent spaces. By gradually incorporating nonlinear factors, the hidden features in the weight vectors are effectively learned, similar to the process of noise addition in a diffusion model. Next, we build a codebook of discrete latent vectors to represent the weight vectors, which corresponds to a clustering operation in latent spaces, with the number of clusters matching the codebook size. Each latent weight vector is then replaced by its closest representation in the codebook. Finally, a decoder is constructed to map the representation vectors back to the original space, aiming to recover the original model weights. The following sections will provide a detailed explanation of the design of these modules.

\subsection{Encoding Weights into Latent Spaces}
The majority of weights in large language models come from the mapping matrices of the linear layers, so we should focus on how to compress these weights. We divide the fitting problem of the entire weight matrix into fitting problems of multiple subvectors. Given a weight matrix $W\in\mathbb{R}^{d_{in}\times d_{out}}$, we divide each row $W_i\in\mathbb{R}^{d_{out}}$ into several subvectors so we can operate at a finer granularity:
\begin{equation}
	W_i = [W_i^1,W_i^2,\cdots,W_i^L],
\end{equation}
where $W_i^L\in\mathbb{R}^{d}$ is the subvector, $L$ is the number of subvectors and $d=d_{out}/L$. For all the weight matrices in a LLM, we can obtain a large number of subvectors, denoted as $S\in\mathbb{R}^{N\times d}$ where $N=d_{in}L$ is the total number of subvectors.

For these weight vectors $S \in \mathbb{R}^{N\times d}$, each of them can be represented as a nonlinear combination of several intrinsic codewords. The nonlinear relationship is hard to explicitly represent, so we propose learning this relationship in a compact latent space. For each subvector $S_i$, a meta encoder $f_e()$ is utilized to transform the subvector into an embedding in latent spaces:
\begin{equation}
	Z_i = f_e(S_i),
\end{equation}
where $Z_i\in\mathbb{R}^{d}$. The meta encoder $f_e()$ can be implemented with a neural network such as Multi-layer Perceptron (MLP). 

Due to the particularity of using weight subvectors as input, we design a simple yet effective norm network to replace the frequently used Layer Normalization (LN) network, called Reshaped Layer Normalization (RLN). As with previous pre-norm works, we perform a norm operation before residual linking to prevent issues such as gradient explosion. After layer-by-layer nonlinear mapping in the encoder network, we will ultimately project the weight vectors into latent spaces for representation in the codebook. It is worth noting that the encoder can be discarded after the end of training. Because we only need to retain the representation vectors in the codebook, the index of each weight vector corresponding to the representation vector, and the decoder.

\paragraph{Reshaped Layer Normalization}
Layer Normalization (LN) is usually used in the Transformer networks to normalize the sequence input embeddings. So it is an intuitive idea to utilize LN networks to process the weight subvectors that are also sequence input. After thorough experimentation, we found that this approach effectively avoids problems such as gradient explosion. However, it also causes a significant difference between the weight vector represented by the codebook and the original vector. Considering that for the input $1 \times d$ sequence (i.e., weight subvector), LN only considers the relationships between $d$ elements in the sequence and ensures that the distribution of these elements is uniform, without considering elements from other input sequences. However, each $1 \times d$ vector is only a part artificially split from the row vectors of the weight matrix, and the elements often do not satisfy a certain distribution relationship or have incomplete semantic associations with each other. Because these elements tend to satisfy a certain distribution within the entire row vectors, we believe that normalizing the original row vectors is more meaningful. Based on the above considerations, we reshape the $1 \times d$ weight vectors back to their original size (i.e. the entire weight row vector) before normalizing. It is equivalent to aligning the vector elements once at the semantic level. After normalization, we then divide the entire row vector into the $1 \times d$ weight subvectors. We name this operation as Reshaped Layer Normalization (RLN). Extensive experiments show that RLN can greatly improve the effect without introducing additional parameter quantities or computational overhead. 

\subsection{Compression in Latent Spaces}
As mentioned above, a codebook is composed of a set of discrete vector sets in latent spaces. We actually cluster weight vectors in latent spaces to obtain $K$ category centers. Each weight vector in latent spaces is represented by the class center closest to it, thus achieving the purpose of compression. Specifically, given weight vectors of a certain layer ${Z} \in \mathbb{R}^{N\times d}$ in latent spaces. At the same time, we initialize $1 \times d$ vectors with the quantity of $K$ in the specific codebook $C \in \mathbb{R}^{K \times d}$. For the weight vector $Z_i$ and codebook $C$, we construct a nearest neighbor mapping. Each latent embedding $Z_i$ is approximated by a codeword (Eq.~\ref{eq:lookup}).
\begin{equation}\label{eq:lookup}
	q(Z'_i=C_k) = \begin{cases} 
		1, & \text{if } k =\arg\min_j \|Z_i-C_j\|_2 \\ 
		0, & \text{otherwise},
	\end{cases}
\end{equation}
where $Z'_i$ is the approximate discrete latent vector of $Z_i$ based on the codebook. This mapping can be represented by an index array $I \in \mathbb{R}^{N}$, which contains $N$ integers. In this way, the weight matrix of a certain layer can be compressed as a codebook and an index array. 

The forward process in Eq.~\ref{eq:lookup} is non-differential, meaning it cannot be optimized by a standard back-propagation algorithm. We approximate the gradient using straight-through estimator~\cite{bengio2013estimating} and directly pass-through the gradient of $Z'$ to $Z$:
\begin{equation}
	\frac{\partial \ell}{Z} = \frac{\partial \ell}{Z'}.
\end{equation}
\begin{figure}[ht]
	\vspace{-1.0em}
	\centering
	\includegraphics[width=0.99\linewidth]{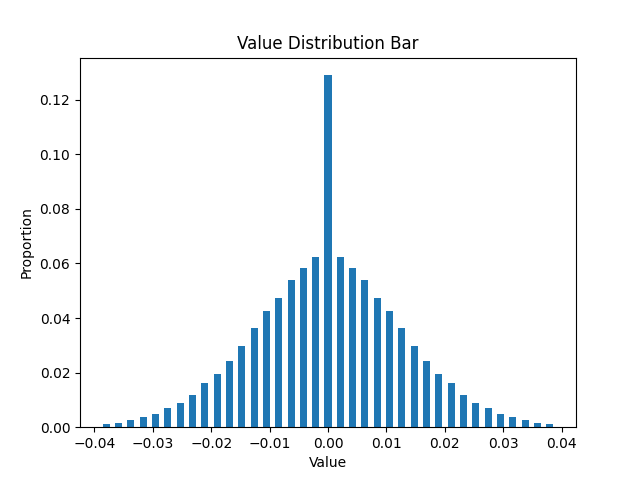}
	\caption{
		We visualize the numerical distribution of values within the 99.9\% range in the value projection weights $W_v$.}
	\label{fig:bar}
\end{figure}

Meanwhile, we conduct experimental statistical analysis of the numerical distribution in the weight vectors and find that, except for a few outliers, most values closely approximate a normal distribution. As shown in Figure~\ref{fig:bar}, taking a projection layer of MLPs as an example, most values are concentrated in a certain interval and approximately follow a normal distribution. In order to approximate the true spatial distribution of vectors and reduce the difficulty of fitting, we use the normal distribution during initialization. We optimize the clustering algorithm using the simplest nearest neighbor algorithm, and only consider the distance between each weight vector and the cluster center during the training process, decoupling it from the subsequent decoding process. Therefore, we utilize Mean Squared Error (MSE) to optimize the codebook by minimizing the Euclidean distance between each weight vector and its representation vector. Specifically, we have $Z'_i$ (Eq.~\ref{eq:lookup}) as the representative vector for the $i$-th latent weight vector $Z_i$, the loss function can be calculated as follows:
\begin{equation}
	MSE =\sum_{i=1}^{N}\|Z_{i}-Z'_i \|^{2}.
	\label{eq:mse}
\end{equation}

\subsection{Weight Reconstruction with Meta Decoders}
As we supposed above, the subvector is a complex nonlinear transformation of an intrinsic codeword. This transformation is difficult to be captured with simple operations like linear fitting, so we propose using a meta decoder to reconstruct the original subvector. The approximated codeword is fed into the meta decoder $f_d()$ for reconstruction:
\begin{equation}
	\hat{S}_i = f_d(Z'_i),
\end{equation}
where $\hat{S}_i\in\mathbb{R}^{d}$ is the reconstructed subvector.

In order to maintain consistency with the structure of the encoder, we samely construct a $m$-layer MLP and use residual links in every layer except for the last layer. Similarly, we use RLN instead of LN before the residual links. The input of the decoder is the representation vector of each weight vector in the latent space. The decoding process is the process of mapping the representation vectors back to the original vector space and expecting to obtain the same output as the input $S$ of the Encoder. We denote the output of the decoder as $\hat{S} \in \mathbb{R}^{N\times d}$ and reduce the difference between $S$ and $\hat{S}$ based on Root Mean Squared Error (RMSE). The specific loss function can be calculated as:
\begin{equation}
	RMSE=\sqrt{\sum_{i=1}^{N}\|S_{i}-\hat{S}_{i} \|^{2}}.
	\label{eq:rmse}
\end{equation}

\subsection{Compression Ratio Analysis}
This section analyzes the compression ratio of the proposed method.
The number of parameters of the original weight matrices is $Nd$. For the compressed version, we only need to store the codebook, the indices, and the meta decoder. Their numbers of parameters are $Kd$, $N$ and $N_{fd}$, respectively. Thus, the compression ratio can be calculated as:
\begin{equation}
	r = \frac{Nd}{Kd+N+N_{fd}}.
\end{equation}
Moreover, if the length of the codebook is $K$, we can store the indices as integers with $\log_2(K)$ bits. We perform half-precision quantization on the codebook. Compared to the original FP32 model, we can calculate the compression ratio as follows:
\begin{equation}
	r = \frac{32Nd}{16Kd+\log_2(K) N+32N_{fd}}.
\end{equation}
Using the Llama 2-7B as the original model for compression, and taking a 3-layer MLP as an example of the meta decoder, its parameter count is 768. We maintain a codebook with $K=2^{15}$ and $d=8$. We have the compression ratio for one $up$ layer of the FFN module:
\begin{equation}
	r = \frac{32\cdot45.1M}{16\cdot 2^{15}\cdot 8 + \log_2(2^{15})\cdot 5.6M + 32\cdot 768} = 16.4
\end{equation}

\subsection{Algorithm Process Overview}
In this section, we use pseudocode to briefly describe our algorithm process.

\setlength{\textfloatsep}{0.3cm}
\begin{algorithm}[ht!]
	\caption{PocketLLM: Weight Compression for LLMs}
	\label{alg:highlvel}
	\small
	\begin{algorithmic}[1]
		\vspace{-1px}\REQUIRE model,calibration\_data,$k$,$d$,$h$
		\FOR{$i = 1, \dots, \texttt{model.num\_layers} $}
		\STATE $f_e,f_d := \texttt{initialize}(h)$ 
		\STATE $C := \texttt{initialize}(k)$ 
		\STATE $\texttt{block := model.get\_block(i)}$\\
		\FOR{$\texttt{layer} \in \texttt{linear\_layers(block)}$} 
		
		\STATE $W := \texttt{layer.weight}$
		\STATE $S := \texttt{split}(W,d)$ 
		\STATE $Z := f_e(S)$
		\STATE $Z' := \texttt{K-means}(Z,C)$ 
		\STATE $\hat{S} := f_d(Z')$
		\STATE $L := RMSE(S,\hat{S})+\lambda MSE(Z,Z')$
		\STATE \texttt{/* After minimizing the loss $L$ */}
		\STATE $\hat{W} := \texttt{merge}(\hat{S},d)$ 
		\STATE $\texttt{layer.weight} := \hat{W}$
		\ENDFOR
		\STATE \texttt{/* save for fine-tuning if necessary */}
		\STATE $\texttt{model} := \texttt{LoRA(model,calibration\_data)}$
		\ENDFOR
	\end{algorithmic}
\end{algorithm}
\vspace{-0.5em}

\begin{table*}[ht!]
	\footnotesize
	\centering
	\setlength\tabcolsep{1.5pt}
	\scalebox{0.896}{
		
		\begin{tabular}{lcc|cccccc}
			\toprule
			\bf{Compress Ratio} & \bf{Method} & \bf{Avg\_bits}  &\bf{WinoGrande$\uparrow$} & \bf{PiQA$\uparrow$} & \bf{HellaSwag$\uparrow$} & \bf{ArcE$\uparrow$} & \bf{ArcC$\uparrow$} & \bf{Avg\_acc$\uparrow$}\\
			\midrule
			
			-- & Llama 2-7B & 32 & 67.25 & 78.45 & 56.69 & 76.01 & 43.03 & 64.29\\
			\midrule
			\multirow{20}{*}{2$\times$$\sim$8$\times$}	
			& LLM-Pruner & 11.20 & 64.25 & 76.01 & - &66.62 & 37.20 & - \\
			& Bolaco & 11.20 & 65.67 & 75.19 & - &69.15 & 38.74 & - \\
			& GPTQ & 4.00 & \underline{68.19} & 76.61 & 55.44 & 66.20 & 36.77 & 60.64 \\
			& SpQR & 3.98  & 66.93 &78.35 & \underline{56.10} & 69.11 & 39.68 & 62.17 \\
			& DuQuant & 4.00  & 63.93 & 75.68 & - & 50.00 & 37.46 & - \\
			& QuIP\# & 4.02   & 66.85 & 77.91 & 55.78 & 68.06 & 39.68 & 61.66 \\
			& AQLM & 4.04  & 67.32 & 78.24 & 55.99 & \underline{70.16} & \underline{41.04} & \underline{62.55} \\
			& VPTQ & 4.01  & 67.10 & 78.10 & 55.00 & 69.00 & 39.70 & 61.98 \\
			& QTIP & 4.01  & 67.10 & \underline{78.40} & - & 68.90 & 40.00 & - \\
			& PocketLLM (ours) & 3.98 & \textbf{69.39} & \textbf{78.54} & \bf{57.45} & \bf{76.18} & \bf{43.17} & \bf{64.95} \\
			\cmidrule{2-9}
			& LLM-Pruner* & 11.20 & 58.72 & 71.93 & - &61.41 & 33.96 & - \\
			& SoLA* & 11.20 & 62.83 & 69.97 & - &59.13 & 32.68 & - \\
			& Bolaco* & 11.20 & 64.80 & 71.44 & - &66.46 & 34.39 & - \\
			& SoBP* & 11.20 & 66.22 & 73.50 & - & 59.81 & 37.63 & - \\
			& Wanda* & 8.00 & \underline{66.61} & - & \underline{46.45} & \underline{69.15} & 35.84 & - \\
			& DuQuant* & 4.00  & 62.12 & \underline{75.24} & - & 51.89 & 36.77 & - \\
			& PocketLLM* (ours) & 3.98  & \textbf{68.19} & \textbf{76.93} & \bf{55.01} & \bf{73.99} & \bf{41.38} & \bf{63.10} \\
			\midrule
			
			\multirow{9}{*}{$\sim10\times$} 
			& GPTQ & 3.00  & 59.19 & 71.49 & 45.21 & 58.46 & 31.06 & 53.08 \\
			& SpQR & 2.98 &  63.54 & 74.81 & 51.85 & 67.42 & 37.71 & 59.07 \\
			& AQLM & 3.04   & 66.93 & 76.88 & 54.12 & 68.06 & 38.40 & 60.88 \\
			& GPTVQ & 3.13  & 67.64 & 77.64 & - &\underline{72.73} & \textbf{43.69} & - \\
			& VPTQ & 3.01  & \textbf{68.00} & 77.30 & \underline{56.00} & 69.10 & 39.30 & \underline{61.72} \\
			& QTIP & 3.01 & 66.90 & \underline{78.10} & - & 68.10 & 38.90 & - \\
			
			& PocketLLM (ours) & 2.98 & \underline{67.40} & \textbf{78.13} & \bf{57.17} & \bf{74.12} & \underline{43.52} & \bf{64.07} \\
			\cmidrule{2-9}
			& PocketLLM* (ours) & 2.98   & \textbf{63.54} & \textbf{75.41} & \bf{51.67} & \bf{69.99} & \bf{37.03} & \bf{59.53} \\
			\midrule
			\multirow{8}{*}{$\sim16\times$} 
			& QuIP\# & 2.02   & 62.43 & 71.38 & 42.94 & 55.56 & 28.84 & 52.23 \\
			& AQLM & 2.02  & \underline{65.67} & 74.76 & 49.55 & 63.68 & 32.76 & 57.28 \\
			& GPTVQ & 2.13 & 60.93 & 70.73 & - &58.08 & 31.48 & - \\
			& VPTQ & 2.02  & 64.33 & 75.19 & \underline{52.08} & 63.80 & 35.24 & \underline{58.13} \\
			& QTIP & 2.00  & 64.70 & \underline{75.90} & - & \underline{67.25} & \underline{35.70} & - \\
			& PocketLLM (ours) & 2.02  & \textbf{67.25} & \textbf{76.71} & \bf{53.24} & \bf{69.07} & \bf{36.77} & \bf{60.61} \\
			\cmidrule{2-9}
			& PocketLLM* (ours) & 2.02 & \textbf{58.72} & \textbf{70.95} & \bf{44.02} & \bf{58.16} & \bf{30.38} & \bf{52.45} \\
			\midrule
			
			\multirow{3}{*}{$\sim20\times$} 
			& PocketLLM (ours) & 1.60  & \textbf{60.30} & \textbf{73.50} & \bf{47.63} & \bf{63.58} & \bf{32.25} & \bf{55.45} \\
			\cmidrule{2-9}
			
			& PocketLLM* (ours) & 1.60   & \textbf{52.01} & \textbf{54.08} & \bf{27.20} & \bf{33.04} & \bf{23.21} & \bf{37.91} \\
			\midrule
		\end{tabular}
	}
\caption{Evaluation of compressed Llama 2-7B models for the accuracy of 5 zero-shot tasks under different compression ratios. The Avg\_acc column is the mean of these accuracies. * means no fine-tuning in the method. \_ represents the second-best result.}
\label{tab:llama2allbit}%
	\vspace{-0.0em}
\end{table*}

\section{Experiments}
We first introduce the implementation details. Then we compare PocketLLM against other approaches under both fine-tuning and non-fine-tuning settings. Next, we show our visual results. Finally, we make a detailed ablation analysis.

\begin{table*}[ht!]
	
	\footnotesize
	\centering
	\setlength\tabcolsep{1.5pt}

	\scalebox{0.89}{
		\begin{tabular}{lcc|cccccc}
			\toprule
			\bf{Compress Ratio} & \bf{Method} & \bf{Avg\_bits}& \bf{WinoGrande$\uparrow$} & \bf{PiQA$\uparrow$} & \bf{HellaSwag$\uparrow$} & \bf{ArcE$\uparrow$} & \bf{ArcC$\uparrow$} & \bf{Avg\_acc$\uparrow$}\\
			\midrule
			
			--& Qwen 3-14B & 32 & 74.35 & 80.58 & 61.86 & 83.54 & 55.80 & 71.23\\
			\midrule
			\multirow{5}{*}{$\sim8\times$} & SmoothQuant & 4.00  & 63.20 & 72.40 & 46.30 &69.60 & 43.40 & 58.98  \\
			& RTN & 4.00  & 65.40 & 76.30 & 53.50 & 76.40 & 51.00 & 64.52 \\
			& AWQ & 4.00  & 71.70 & \underline{79.90} & \underline{60.60} & \underline{82.20} & \underline{54.20} &  69.72\\
			& GPTQ & 4.00  & \underline{73.20} & 79.70 & \underline{60.60} & 81.50 & 54.00 &  \underline{69.80}\\
			& PocketLLM (ours) & 3.98  & \textbf{74.90} & \textbf{80.14} & \textbf{60.64} & \textbf{84.85}& \textbf{55.97}   & \textbf{71.30} \\
			\midrule
			\multirow{2}{*}{$\sim10\times$} 
			& AWQ & 3.00  & \underline{62.70} & \underline{76.00} & \underline{55.90} & \underline{73.00} & \underline{44.60} &  \underline{62.44}\\
			& PocketLLM (ours) & 3.02  & \textbf{74.20} & \textbf{79.49} & \textbf{57.72} & \textbf{84.22}& \textbf{55.25}   & \textbf{70.19} \\
			\midrule
		\end{tabular}
	}
\caption{Evaluation of compressed Qwen 3-14B models for the accuracy on 5 zero-shot tasks.}
\label{tab:qwen}%
	\vspace{-0.5em}
\end{table*}

\subsection{Implement Details}
We apply the PocketLLM algorithm based on the Llama 2-7B and Qwen 3-14B model provided by Huggingface~\cite{touvron2023llama2openfoundation,qwen3}. We use GELU activation function~\cite{hendrycks2016gelu} in MLP to increase the nonlinearity of the network, which helps to perceive potential differences between vectors. We construct a $3$-layer MLP network for both encoder and decoder, and each layer except for the first layer uses residual links to improve performance. We set the value of $k$ (the size of the codebook) and $d$ (the length of weight vector) to $(d,k)=(4,2^{15})$, $(d,k)=(4,2^{12})$, $(d,k)=(8,2^{15})$, $(d,k)=(8,2^{12})$, corresponding to $8\times$, $10\times$, $16\times$, and $20\times$ compression experiments. To fairly compare with recent methods, we conducted experiments with and without fine-tuning simultaneously. Different from some methods using block-wise or layer-wise fine-tuning step-by-step during the compression~\cite{egiazarian2024extreme}. We only use the standard LoRA algorithm for fine-tuning once a time after compression. Specifically, we set rank and alpha as 32 and 64, respectively. Other parameters are set to the default value. We use the same calibration datasets (redpajama~\cite{weber2024redpajama},alpaca~\cite{alpaca}) as recent methods for fine-tuning~\cite{deng2024llmcodebook,egiazarian2024extreme}. 

\subsection{Comparisons with State-of-the-art Methods}
In this section, we compare the PocketLLM with other methods on 7 benchmarks. Specifically, we report the accuracy on five zero-shot tasks, i.e., WinoGrande~\cite{DBLP:journals/cacm/winogrande2021}, PiQA~\cite{tata2003piqa}, HellaSwag~\cite{DBLP:conf/acl/hellaswag2019}, ARC-easy (Arc-E), and ARC-challenge (Arc-C)~\cite{boratko2018systematic}, and report the perplexity on WikiText-2~\citep{wikitext103} and C4~\citep{C4} datasets. As with the previous work, the calculation of the $average$ $bits$ only takes quantized weights into account.

In the Table~\ref{tab:llama2allbit}, we compare the accuracy of PocketLLM and other methods on five zero-shot tasks. According to the results, we can find that PocketLLM achieves better results regardless of fine-tuning or not at different compression ratios. Especially under the $8\times$ (compared to the float32 model) compression settings, PocketLLM outperforms other methods by a large margin. Even compared to the uncompressed model, PockerLLM without fine-tuning still has a comparable performance. Meanwhile, under the $10\times$ compression settings, PocketLLM without fine-tuning has a certain loss in accuracy. However, with just a simple fine-tuning, the performance can be well restored. Under the $16\times$ compression settings, the performance loss of PocketLLM with fine-tuning is still acceptable. When the compression ratio reaches $20\times$, PocketLLM is still able to greatly restore the performance through fine-tuning. 


In the Table~\ref{tab:qwen}, we also extend PocketLLM on the base model Qwen 3-14B~\cite{qwen3} for $8\times$ and $10\times$ compression ratios, respectively. According to the results, we can find that the precision advantage of PocketLLM is significant. Even if compressed $10\times$ times, the loss of accuracy is still negligible. More compressing results about Llama 3-8B and Llama 1-7B can be found in supplementary materials. 

In the Table~\ref{tab:llama2ppl}, we also compare the perplexity of PocketLLM and other methods. According to the results, we can find that PocketLLM achieves comparable performance to AQLM, QTIP, and surpass all the other methods. As for the accuracy gap, we analyze that it is due to our insufficient calibration fine-tuning. The rougher precision recovery approach may not be suitable for the perplexity metrics.

%
\subsection{Ablation Study}
In this section, we mainly make ablation studies based on quantized Llama 2-7B without fine-tuning under $8\times$ compression. We first analyze the difficulty of compressing for different types of layers. Then, we analyze the impact of MLP quantity on the reconstruction loss. Next, we analyze the impact of codebook size. Finally, the effectiveness of RLN and codebook initialization is analyzed. 

In the Table~\ref{tab:llamalayer}, in addition to the standard 7 layers, we also investigate the losses caused by compressing all the attention layers ($q, k, v, o$) and all the FFN layers ($gate, up, down$), respectively. Unlike previous experiments, we also add the MMLU~\cite{mmlu} dataset to observe the loss of different layers in difficult scenarios. It can effectively prevent fluctuations on simple datasets from affecting conclusions. According to the results, we can observe interesting phenomena. Firstly, although a single FFN layer has more parameters, its impact on the loss is not much higher than that of a single attention layer. Second, compresing all the attention layers will bring apparent accuracy loss that is almost equivalent to compressing all the FFN layers. Therefore, we can conclude that even if the number of parameters occupied by attention layers is less than one-third, it is still important. Perhaps this can inspire us to design LLMs with limited parameters (smaller than 2B), and we may be able to lose some parameters in FFN modules to ensure the parameters in attention layers.
\begin{table}[ht!]
	\footnotesize
	\centering
	\setlength\tabcolsep{3pt}
	\scalebox{0.89}{
		\begin{tabular}{lcc|cc}
			\toprule
			\bf{Ratio} & \bf{Method} & \bf{Avg\_bits} & \bf{WikiText-2$\downarrow$} & \bf{C4$\downarrow$} \\
			\midrule
			-- &Llama 2-7B & 32 & 5.12 & 6.63\\
			\midrule
			\multirow{10}{*}{$\sim8\times$} 
			& GPTQ & 4.00 & 5.49 & 7.20 \\
			& SpQR & 3.98  & 5.28 & 6.87  \\
			& QuIP\# & 4.02  & 5.29 & 6.86 \\
			& DuQuant & 4.00  & 6.08 & 7.79 \\
			& AQLM & 4.04 & \underline{5.21} & \underline{6.75} \\
			& VPTQ & 4.01 & 5.64 & 7.13 \\
			& QTIP & 4.01 & \textbf{5.17} & \textbf{6.69} \\
			& PocketLLM & 3.98  & 5.27 & 6.86 \\
			\midrule
			& SoBP* &11.20 & 8.64 & - \\
			& Wanda* & 8.00 & 6.42 & - \\
			& DuQuant* & 4.01 & \underline{6.28} & \underline{7.90} \\
			& PocketLLM & 3.98  & \textbf{5.74} & \textbf{7.49} \\
			\midrule
		\end{tabular}
	}
\caption{The perplexity of quantized Llama 2-7B models with weight updation under $8\times$ compression ratio. \_ represents the second-best result.}\label{tab:llama2ppl}%
	\vspace{-0.5em}
\end{table}

In the Table~\ref{tab:mlpnums}, we explore the optimal number of model layers for the encoder and decoder. We denote $vq\_loss$ and $mse\_loss$ as $vq$ and $mse$, respectively. The former can be used to characterize the distribution of latent vectors in the codebook. The smaller $vq$ indicates a more uniform distribution and better representation of latent vectors in the codebook. The latter represents the overall loss of the reconstructed vectors. Besides, we also calculate the sum of the top 100 values of $mse\_loss$, denoted as $mse\_top100$. It can effectively assist us in observing the outlier situation of the reconstructed vectors. We can observe that as the number of model layers increases, both $mse\_loss$ and $vq\_loss$ decrease significantly. However, as the number of layers increases further, $vq\_loss$ significantly increases. We believe that with the further growth of the MLP layer, too many nonlinear factors have been added, making it hard to learn the hidden features of the original vectors well. 

\begin{table}[ht!]
	\footnotesize
	\centering
	\setlength\tabcolsep{3pt}
	
	\scalebox{0.89}{
		\begin{tabular}{lcc|cc}
			\toprule
			\bf{Method}&\bf{layer} & \bf{rate} &\bf{MMLU (5)}  & \bf{HellaSwag (0)} \\
			\midrule
			Llama 2-7B & -- & -- &45.34 & 56.69 \\
			\midrule
			\multirow{11}{*}{PocketLLM}&q & 8.3\% &45.84 & 56.43 \\
			&k & 8.3\% &46.10 & 56.12 \\
			&q,k & 16.6\% &45.72 & 56.35 \\
			&v & 8.3\% &45.66 & 57.23 \\
			&o & 8.3\% &45.32 & 55.82 \\
			&q,k,v,o & 33.2\% &44.14 & 55.93 \\
			&gate & 22.3\% &45.71 & 56.63 \\
			&up & 22.3\% &45.63& 55.89 \\
			&down & 22.3\% &45.62& 54.55 \\
			&gate,up,down & 66.8\% &43.53& 55.74 \\
			&all & 100\% &41.50& 55.01 \\
			\midrule
		\end{tabular}
	}
\caption{The ablation study on Llama 2-7B models that compress different type of layers for $8\times$ compression ratio. }
\label{tab:llamalayer}%
	\vspace{-0.5em}
\end{table}

\begin{table}[ht]
	\footnotesize
	\centering
	\setlength\tabcolsep{5pt}
	
	\scalebox{0.89}{
		\begin{tabular}{c|ccc}
			\toprule
			\bf{mlp\_layers}  &\bf{vq}  & \bf{mse} & \bf{mse\_top100}  \\
			\midrule
			1& 4.67 & 4.8e-5 & 1.26  \\
			2&0.46 & 1.1e-5& 0.12 \\
			3&0.06 & 8.9e-6 & 0.08 \\
			5&0.20 & 9.2e-6 & 0.09\\
			\midrule
		\end{tabular}
	}
\caption{The ablation study on Llama 2-7B models that construct MLP layers in encoder and decoder.}
\label{tab:mlpnums}%
	\vspace{-0.5em}
\end{table}

\begin{table}[ht]
	\footnotesize
	\centering
	\setlength\tabcolsep{5pt}
	
	\scalebox{0.89}{
		\begin{tabular}{c|ccc}
			\toprule
			\bf{codebook\_size}  &\bf{vq}  & \bf{mse} & \bf{mse\_top100}  \\
			\midrule
			256& 5.54 & 8.6e-5 & 0.36  \\
			1024&3.12 & 4.2e-5 & 0.22 \\
			4096&2.41 & 2.2e-5 & 0.16 \\
			16384&0.38 & 1.3e-5 & 0.10\\
			32768&0.06 & 8.9e-6 & 0.08\\
			\midrule
		\end{tabular}
	}
\caption{The ablation study on Llama 2-7B models with different codebook size.}
\label{tab:codebooknums}%
	\vspace{-0.5em}
\end{table}

In the Table~\ref{tab:codebooknums}, we observe the impact of codebooks with different sizes on the results. As the size increases, the loss will gradually decrease. When the size of the codebook is smaller than 4096, reducing the size will significantly increase the losses. While the size is greater than 32768, the downward trend of losses will slow down.

In the Table~\ref{tab:module}, we analyze the effectiveness of RLN and codebook initialization. According to the results, we can find that RLN can effectively reduce the $mse\_loss$ and improve the reconstruction effect. Codebook initialization can significantly reduce the $vq\_loss$ and relieve the outlier situation.

\begin{table}[ht!]
	\footnotesize
	\centering
	\setlength\tabcolsep{5pt}
	
	\scalebox{0.89}{
		\begin{tabular}{cc|ccc}
			\toprule
			\bf{RLN}  &\bf{initial}   &\bf{vq} & \bf{mse} & \bf{mse\_top100}  \\
			\midrule
			&  &  8.51 & 4.5e-5 & 0.89  \\
			&$\surd$ & 2.36 & 2.2e-5 & 0.36 \\
			$\surd$ &  & 1.23 & 9.2e-6 & 0.25 \\
			$\surd$ &$\surd$ &0.06 & 8.9e-6 & 0.08\\
			\midrule
		\end{tabular}
	}
\caption{The ablation study on the proposed RLN module and codebook initialization.}
\label{tab:module}%
	\vspace{-0.5em}
\end{table}

\section{Visualization results}
\begin{figure}[ht!]
	\centering
	\includegraphics[width=0.77\linewidth]{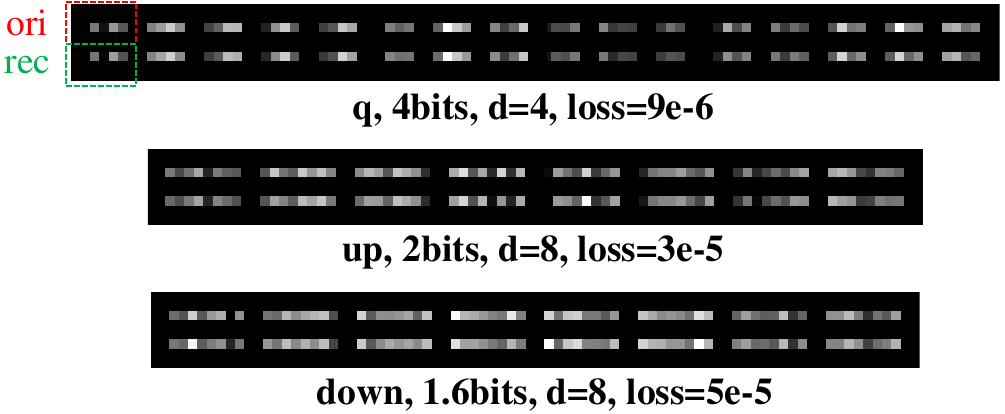}
	\caption{The reconstructed results of $q$, $up$, $down$ weight vectors correspond to the compression ratios of $8\times$ (4 bits), $16\times$ (2 bits), and $20\times$ (1.6 bits), respectively.}
	\label{fig:visualize}
	\vspace{-1.0em}
\end{figure}

As shown in Figure~\ref{fig:visualize}, we visualize the reconstruction results of $q,up,down$ weight vectors for different compression ratios. We visualize $1\times4$ weight vectors with the quantity of 16 for  $q$ layers. We visualize $1\times8$ weight vectors with the quantity of 8 for $up$ and $down$ layers. As the compression ratio increases, the difference between the original vector and the recovered vector gradually becomes apparent in the details but still remains overall consistent.

\section{Conclusion}
In this paper, we propose PocketLLM, a method for compressing the weights of LLMs at extreme compression ratios. By employing encoder-decoder meta-networks, PocketLLM effectively projects the original weights into a discrete latent space, incorporating nonlinear factors for better representation. It then learns a compact codebook with strong representational power, mapping the codebook's representative vectors back to the original space. This approach significantly improves upon existing low-bit compression techniques. At \(8 \times\) compression, PocketLLM without fine-tuning, outperforms current fine-tuning methods and achieves performance comparable to the dense model. At \(10 \times\) compression, PocketLLM with fine-tuning maintains the dense model's accuracy, and even at \(20 \times\) compression, simple fine-tuning restores accuracy.
\bibliography{aaai2026}

\end{document}